\documentclass[10pt,twocolumn,letterpaper]{article}

\usepackage{cvpr}

\usepackage{times}
\usepackage{epsfig}
\usepackage{graphicx}
\usepackage{amsmath}
\usepackage{amssymb}

\usepackage{graphicx,subcaption}

\usepackage[breaklinks=true,bookmarks=false]{hyperref}

\usepackage{array}
\newcolumntype{P}[1]{>{\centering\arraybackslash}p{#1}}
\newcolumntype{M}[1]{>{\centering\arraybackslash}m{#1}}

\cvprfinalcopy 

\def\cvprPaperID{21} 

\ifcvprfinal\fi

\begin{document}
	
\title{Dropping Pixels for Adversarial Robustness}

\author{Hossein Hosseini \qquad Sreeram Kannan \qquad Radha Poovendran\\
	Department of Electrical Engineering, University of Washington, Seattle, WA \\
	{\small \{hosseinh, ksreeram, rp3\}@uw.edu}\\
}

\maketitle

\begin{abstract}
Deep neural networks are vulnerable against adversarial examples. In this paper, we propose to train and test the networks with randomly subsampled images with high drop rates. We show that this approach significantly improves robustness against adversarial examples in all cases of bounded $L_0$, $L_2$ and $L_{\infty}$ perturbations, while reducing the standard accuracy by a small value. 
We argue that subsampling pixels can be thought to provide a set of robust features for the input image and, thus, improves robustness without performing adversarial training.

\end{abstract}

\section{Introduction}

Deep neural networks are known to be vulnerable against adversarial examples, inputs that are intentionally designed to cause the model to make a mistake~\cite{Goodfellow2017openai}. One particular type of adversarial examples for image classifiers is slightly perturbed images that are misclassified by the model, but are recognizable to humans~\cite{biggio2013evasion, szegedy2013intriguing}. 
Such adversarial images are typically generated by adding a small perturbation with bounded $L_0$, $L_2$ or $L_{\infty}$ norm to legitimate inputs~\cite{carlini2017towards}.

Several methods have been proposed for defending against adversarial examples, but later broken using adaptive iterative attacks~\cite{carlini2017adversarial,athalye2018obfuscated}. 
The state-of-the-art defense against adversarial examples (with bounded $L_{\infty}$ perturbation) is adversarial training, which iteratively generates adversarial examples and trains the model to classify them correctly~\cite{goodfellow2014explaining,madry2017towards}. 
This approach, however, significantly slows down the training process and does not properly scale to large datasets~\cite{kannan2018adversarial}.

\begin{figure}[t]
	\centering
	\begin{subfigure}[t]{0.11\textwidth}
		\centering
		\includegraphics[width=1\linewidth]{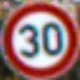}
	\end{subfigure}\hspace{0.0cm}
	\begin{subfigure}[t]{0.11\textwidth}
		\centering
		\includegraphics[width=1\linewidth]{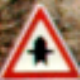}
	\end{subfigure}\hspace{0.0cm}
	\begin{subfigure}[t]{0.11\textwidth}
		\centering
		\includegraphics[width=1\linewidth]{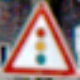}
	\end{subfigure}\hspace{0.0cm}
	\begin{subfigure}[t]{0.11\textwidth}
		\centering
		\includegraphics[width=1\linewidth]{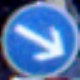}
	\end{subfigure}\\\vspace{0.1cm}
	\begin{subfigure}[t]{0.11\textwidth}
		\centering
		\includegraphics[width=1\linewidth]{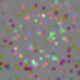}
	\end{subfigure}\hspace{0.0cm}
	\begin{subfigure}[t]{0.11\textwidth}
		\centering
		\includegraphics[width=1\linewidth]{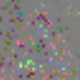}
	\end{subfigure}\hspace{0.0cm}
	\begin{subfigure}[t]{0.11\textwidth}
		\centering
		\includegraphics[width=1\linewidth]{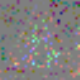}
	\end{subfigure}\hspace{0.0cm}
	\begin{subfigure}[t]{0.11\textwidth}
		\centering
		\includegraphics[width=1\linewidth]{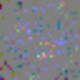}
	\end{subfigure}\\\vspace{0.1cm}
	\begin{subfigure}[t]{0.11\textwidth}
		\centering
		\includegraphics[width=1\linewidth]{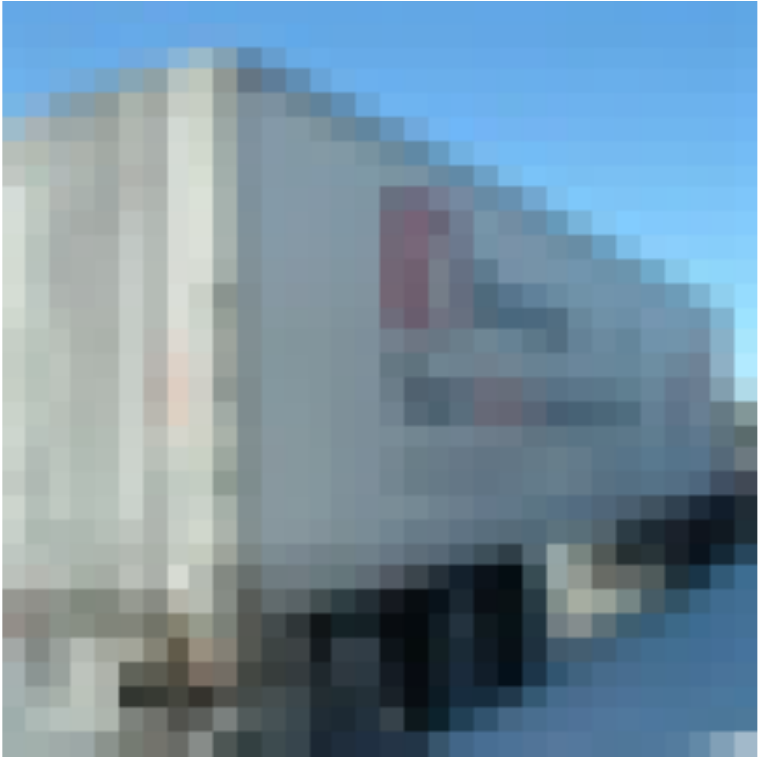}
	\end{subfigure}\hspace{0.0cm}
	\begin{subfigure}[t]{0.11\textwidth}
		\centering
		\includegraphics[width=1\linewidth]{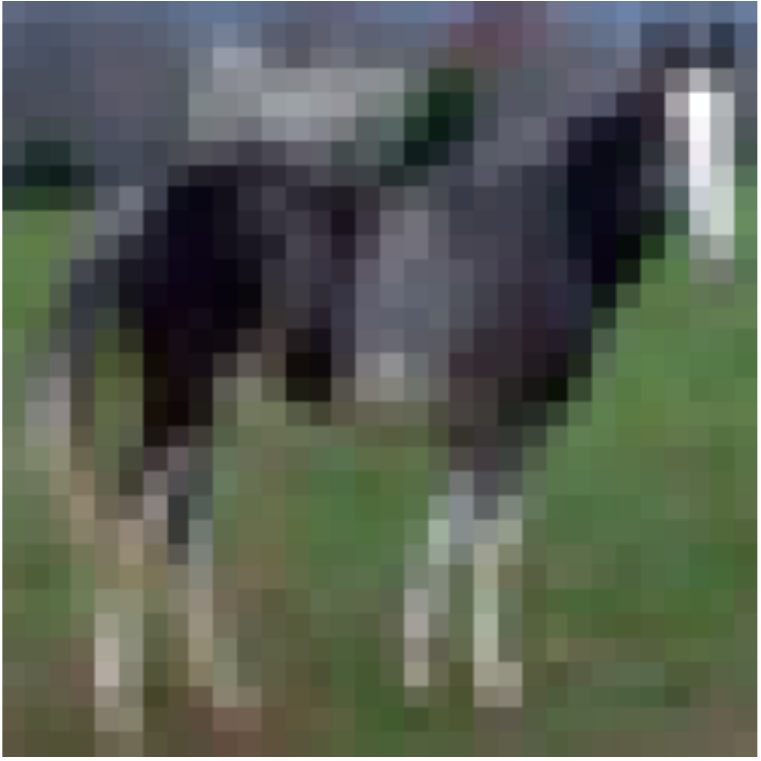}
	\end{subfigure}\hspace{0.0cm}
	\begin{subfigure}[t]{0.11\textwidth}
		\centering
		\includegraphics[width=1\linewidth]{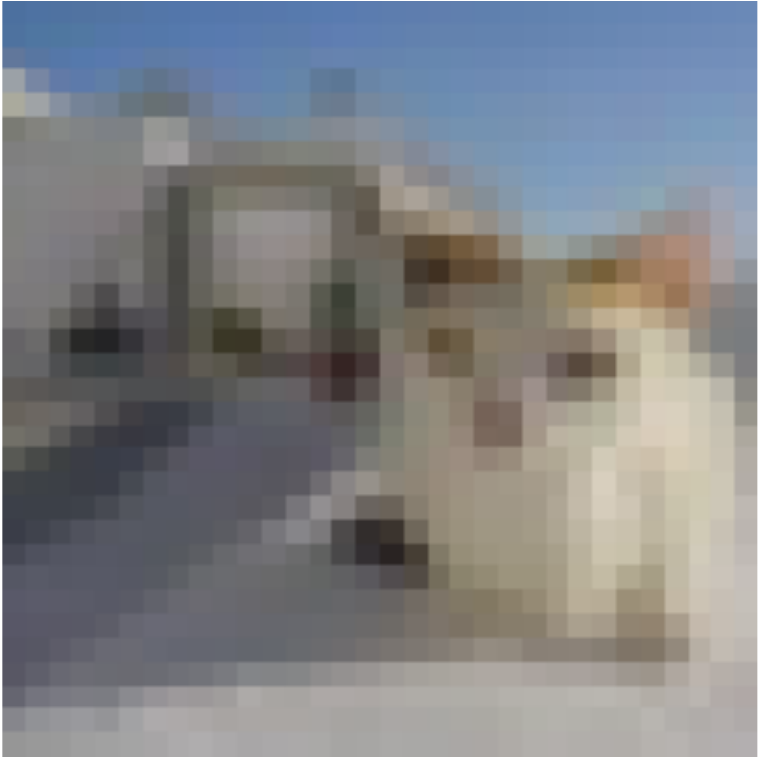}
	\end{subfigure}\hspace{0.0cm}
	\begin{subfigure}[t]{0.11\textwidth}
		\centering
		\includegraphics[width=1\linewidth]{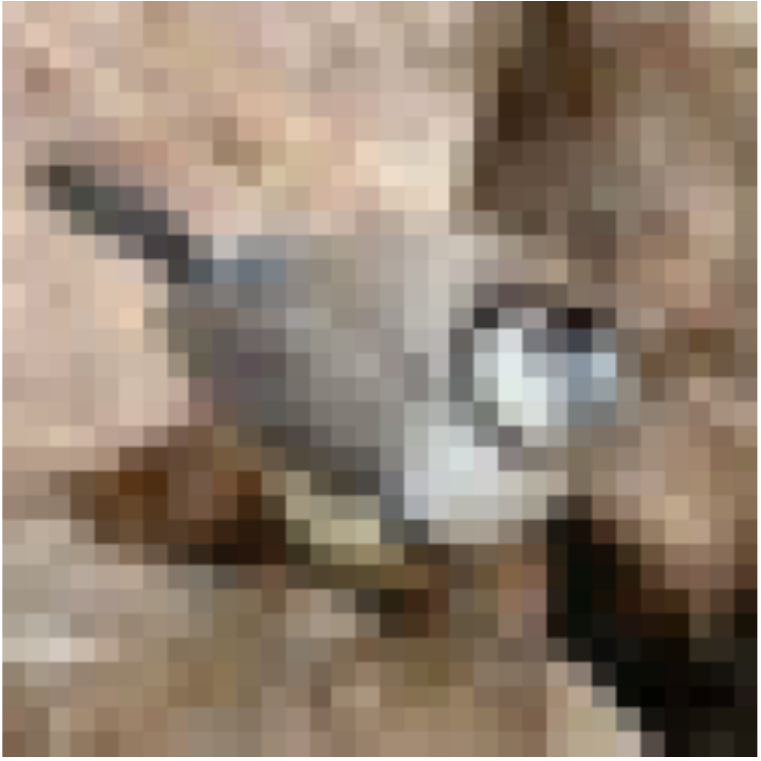}
	\end{subfigure}\\\vspace{0.1cm}
	\begin{subfigure}[t]{0.11\textwidth}
		\centering
		\includegraphics[width=1\linewidth]{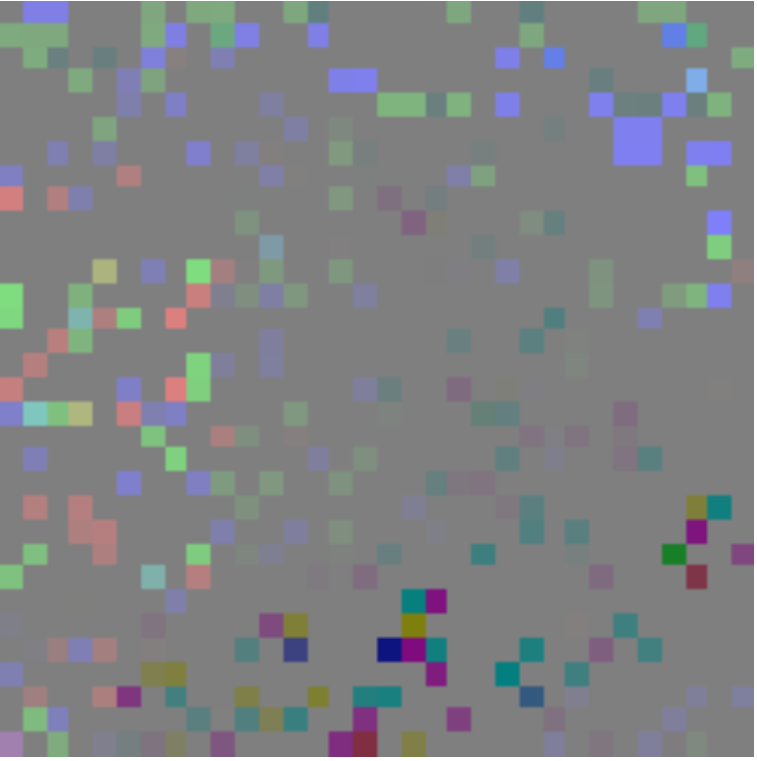}
	\end{subfigure}\hspace{0.0cm}
	\begin{subfigure}[t]{0.11\textwidth}
		\centering
		\includegraphics[width=1\linewidth]{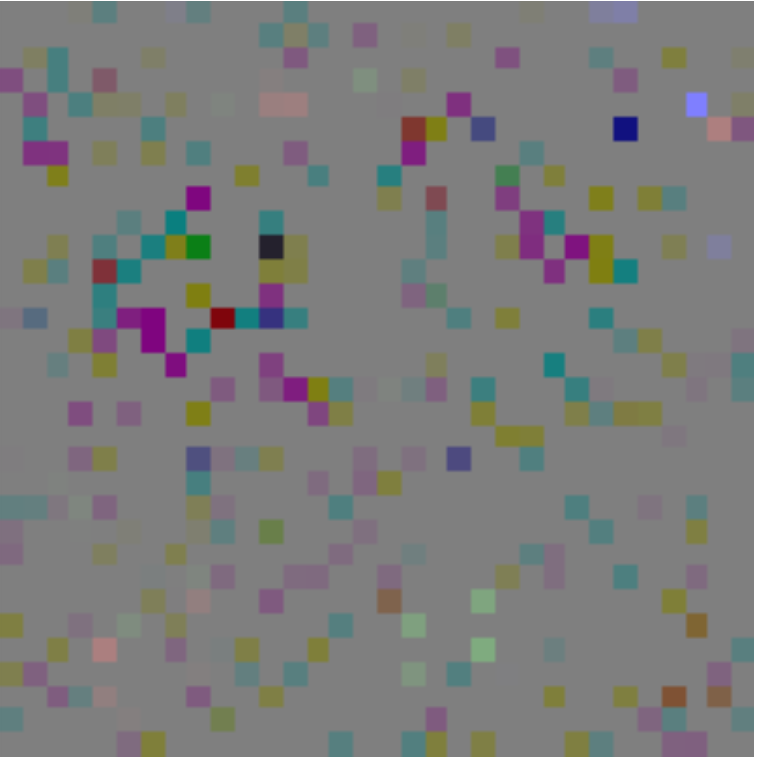}
	\end{subfigure}\hspace{0.0cm}
	\begin{subfigure}[t]{0.11\textwidth}
		\centering
		\includegraphics[width=1\linewidth]{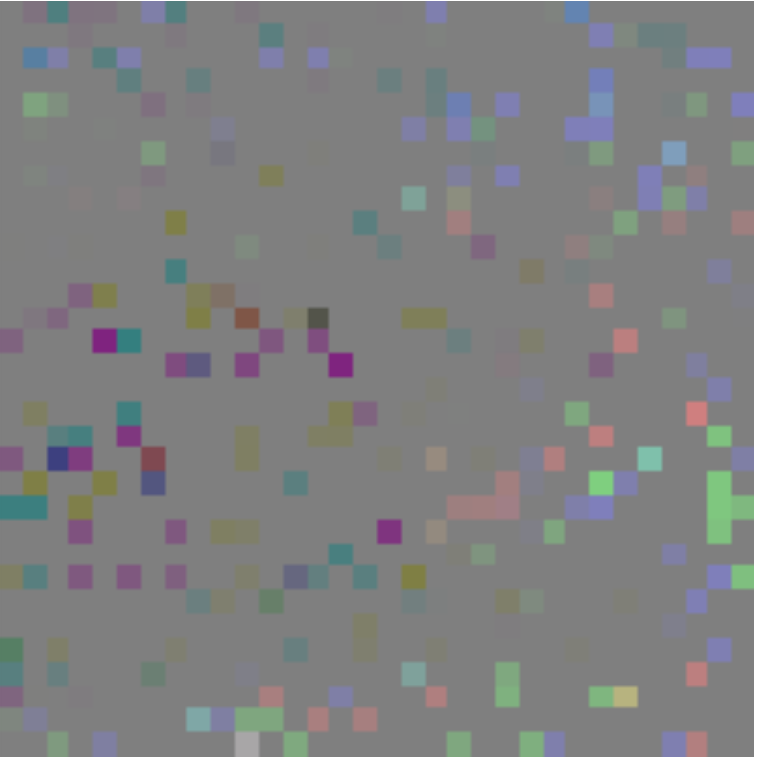}
	\end{subfigure}\hspace{0.0cm}
	\begin{subfigure}[t]{0.11\textwidth}
		\centering
		\includegraphics[width=1\linewidth]{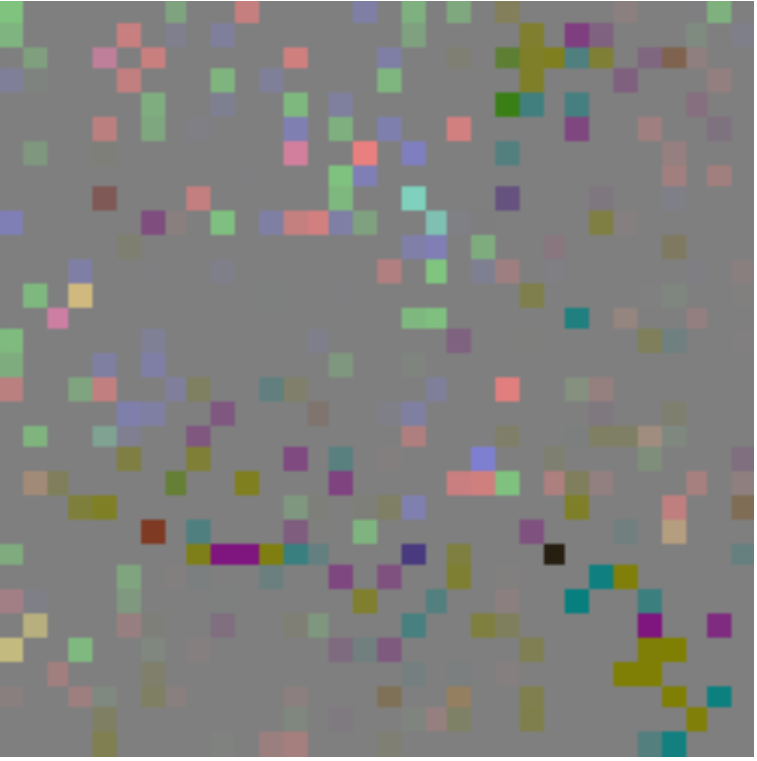}
	\end{subfigure}	
	\caption{Examples of original and subsampled images with drop rate of $90\%$. (First and Third Rows) Images from GTSRB and CIFAR10 datasets, respectively, (Second and Fourth Rows) Corresponding subsampled images. 
		The accuracy on subsampled images reduces by about $4\%$ and $11\%$ compared to original images for GTSRB and CIFAR10 datasets, respectively.}
	\label{fig:rand}
\end{figure}

Adversarial training is shown to improve robustness at the cost of reducing accuracy~\cite{madry2017towards}. In~\cite{tsipras2018there}, Tsipras et al., argued that the trade-off between adversarial robustness and standard generalization is a fundamental property of machine learning classifiers. They analyzed a binary classification problem and showed that the reduction in standard accuracy is due to the tendency of adversarially trained models to assign non-zero weights only to a small number of strongly-correlated or ``robust'' features. That is, such networks discard the weakly-correlated (non-robust) features that could potentially lead to better standard generalization.

In this paper, we investigate how this insight could be used to train robust classifiers without performing adversarial training. 
In natural image classifiers, it is not possible to identify a fixed set of 
robust features in pixel domain due to the position invariance of objects. 
As a result, the set of robust features would be different for each image. 
To adapt the idea of selecting robust features to natural images, we use a slightly different notion of robust features as features that are strongly-correlated with output {\it given all other robust features}. In other words, instead of selecting features that are each highly correlated with output, we select the {\it set} of features that has the highest correlation.

Image data contain high redundancy due to the strong correlation between neighboring pixels, i.e., it is possible to restore images even when a large fraction of pixels is removed~\cite{elad2005simultaneous,hosseini2013fast}. 
Therefore, conditioned on that a pixel is selected, its surrounding pixels are weakly-correlated with output, because they significantly overlap in content with the center pixel and removing them will not cause much reduction in accuracy. 
Hence, one straightforward approach to construct robust features is by downsampling image pixels. Since farther pixels have smaller correlation, they non-trivially contribute to model's prediction and, thus, are considered to be robust features.

We propose to perform random (nonuniform) sampling in order to improve both accuracy and robustness. 
Random subsampling of pixels improves standard generalization because the model will be trained with different subsets of pixels of each image. Also, at inference time, the accuracy can be improved by averaging the prediction over multiple sampling patterns. Moreover, since the randomness is not known to the adversary, it further mitigates the attack success rate. Randomly dropping pixels is suited to defend against adversarial examples with bounded $L_0$ perturbation, since the model learns to recognize objects from images with missing pixels. Nevertheless, we show that it provides robustness against adversarial examples with bounded $L_2$ and $L_{\infty}$ perturbation as well. 

In this paper, we present our preliminary work and results on using random subsampling for adversarial robustness. 
Our contributions are summarized in the following. 

\begin{itemize}
	\item We show image classifiers can be trained with inputs with reduced redundancy, through random subsampling of pixels, without significant reduction in accuracy. We show that the best results are obtained when the model is trained with subsampled images with drop rates chosen randomly in $[0,1]$. 
	\item We apply the interpretability methods on models trained with subsampled images and argue that such approaches cannot explain how the model recognizes images from few pixels. We also visualize convolutional filters of the first layer of the network and show that, in this respect, the model behaves similar to a network trained using adversarial training. 
	
	\item We evaluate adversarial robustness of the models trained with random subsamled images. Experiments are performed on GTSRB and CIFAR10 datasets and with projected gradient descent (PGD)
	attack~\cite{kurakin2016adversarial}. We show that training with subsampled images with drop rates chosen randomly in $[0,1]$ improves the robustness against adversarial examples in all cases of bounded $L_0$, $L_2$ and $L_{\infty}$ perturbation. 
\end{itemize}

\begin{figure}[t]
	\centering
	\begin{subfigure}[t]{0.48\textwidth}
		\centering
		\includegraphics[width=0.85\linewidth]{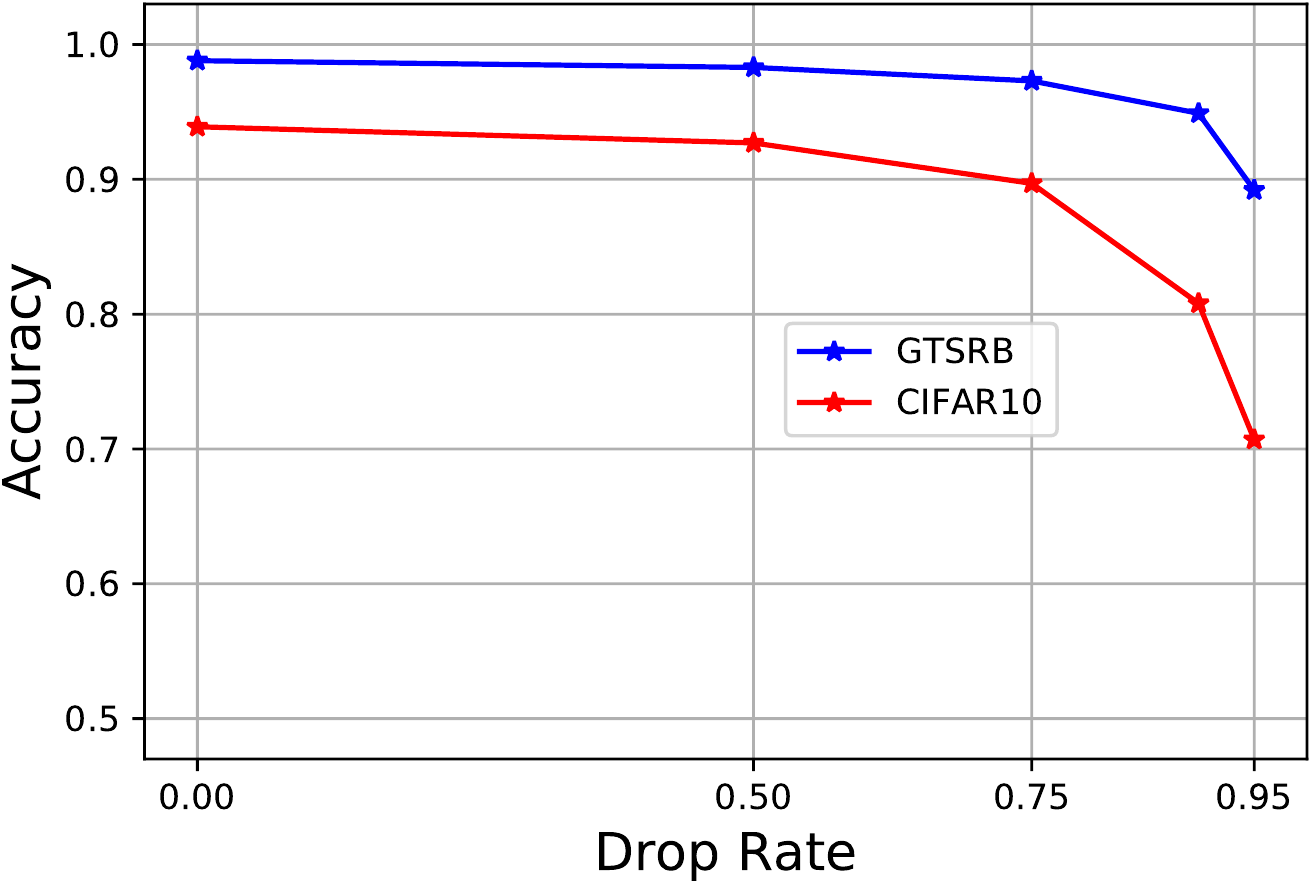}
	\end{subfigure}
	\caption{\small Accuracy of models trained with subsampled images with different drop rates. We used ResNet-20 and ResNet-110 for GTSRB and CIFAR10, respectively. Dropping pixels at a higher rate results in lower accuracy. However, even at very high drop rates, the accuracy remains high.}\label{fig:down}
\end{figure}

\section{Training with Subsampling Pixels}

Natural images are high-dimensional data with high redundancy due to the strong correlation between neighboring pixels. Hence, when training image classifiers, we can potentially reduce the redundancy without significantly reducing the standard accuracy. One approach for reducing redundancy is randomly dropping pixels at a high rate. 
In the following, we provide the results of training and testing models with  images with missing pixels. 

Let $X\in[-1,1]^{d\times d\times 3}$ be a color image. 
Let $M\in\{0,1\}^{d\times d\times 3}$ be a mask with the same size as $X$, where $M_{i,j,k}$ is a Bernoulli random variable with mean $1-r$, i.e., elements of $M$ are equal to $0$ with probability $r$ and equal to $1$ otherwise.  
We generate the subsampled image $X'$ as $X'= M\circ X$, where $\circ$ denotes Hadamard (element-wise) multiplication. 
Figure~\ref{fig:rand} shows samples of original images of German Traffic Sign Recognition Benchmark (GTSRB)~\cite{stallkamp2012man} and CIFAR10 dataset~\cite{krizhevsky2009learning} and their corresponding subsampled images with drop rate of $90\%$. 

\subsection{Experimental Results}

We use ResNet-20 and ResNet-110 architectures~\cite{he2016deep} for GTSRB and CIFAR10 datasets, respectively. The models are trained and tested with subsampled images. During training, the mask is chosen randomly and differently for each image and at each epoch. 
Figure~\ref{fig:down} shows accuracy of models trained with images with different drop rates. As expected, dropping pixels at a higher rate results in lower accuracy. However, even at very high drop rates, the accuracy remains high. 
Specifically, compared to standard training, at drop rate of $90\%$ the accuracy reduces by only about $4\%$ and $13\%$ for GTSRB and CIFAR10 datasets, respectively.

We also observed that deeper networks perform better. Table~\ref{table:deep} shows the accuracy of ResNet models with different depths on CIFAR10 images with drop rate of $90\%$. As can be seen, ResNet-110 provides about $2\%$ and $6\%$ higher accuracy compared to ResNet-56 and ResNet-20, respectively. Moreover, the model achieves best results when the drop rate of each image is chosen randomly between $0\%$ and $100\%$ at each epoch. 

\bgroup
\def\arraystretch{1.1}
\begin{table}[t]
	\vspace{0.cm}
	\centering
	\caption{\small Results on CIFAR10 dataset. 
		In experiment 1, model is trained and tested with original images. 
		In experiment 2, models are trained and tested with subsampled images with drop rate of $90\%$. In experiment 3, model is trained with subsampled images with drop rates chosen uniformly in $[0,1]$ and tested on subsampled images with drop rate of $90\%$.}
	\begin{tabular}{|c|c|}
		\hline 
		{\bf \small Model} & {\bf \small Accuracy}\\
		\hline
		\hline
		{\small Experiment 1 (ResNet-110)} & $93.9\%$ \\
		\hline
		{\small Experiment 2 (ResNet-20)} & $75.2\%$ \\
		\hline
		{\small Experiment 2 (ResNet-56)} & $79.4\%$ \\
		\hline
		{\small Experiment 2 (ResNet-110)} & $81.1\%$ \\
		\hline
		{\small Experiment 3 (ResNet-110)} & $83.0\%$ \\
		\hline
	\end{tabular}
	\label{table:deep}
\end{table}
\egroup

\subsection{Interpretability Analysis}

In recent years, several ``post-hoc'' methods have been proposed for interpreting the predictions of deep convolutional neural networks~\cite{simonyan2013deep,shrikumar2016not,selvaraju2016grad,smilkov2017smoothgrad}. 
Such methods typically identify input dimensions that the output is most sensitive to. 
Let $X$ be the input image, $F$ be the classifier and $E$ be the explanation function that maps inputs to objects of the same shape. 
Most explanation methods are based on some form of the gradient of the classifier function with respect to input~\cite{adebayo2018sanity}. 
In our analysis, we use magnitude of gradient as the explanation map, i.e., $E(X)=|\partial F(X)/\partial X|$~\cite{simonyan2013deep}. 

%
%

\begin{figure}[t]
	\centering
	\begin{subfigure}[t]{0.5\textwidth}
		\centering
		\hspace{0.5cm}	\includegraphics[width=0.88\linewidth]{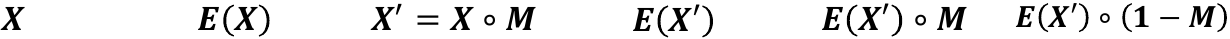}
	\end{subfigure}	\\\vspace{0.15cm}
	\begin{subfigure}[t]{0.5\textwidth}
		\centering
		\includegraphics[width=0.13\linewidth]{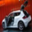}
		\hspace{0.0cm}
		\includegraphics[width=0.13\linewidth]{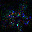}
		\hspace{0.cm}
		\includegraphics[width=0.13\linewidth]{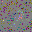}
		\hspace{0.cm}
		\includegraphics[width=0.13\linewidth]{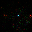}
		\hspace{0.cm}
		\includegraphics[width=0.13\linewidth]{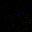}
		\hspace{0.cm}
		\includegraphics[width=0.13\linewidth]{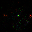}
		\\\vspace{0.15cm}
		\includegraphics[width=0.13\linewidth]{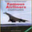}
		\hspace{0.0cm}
		\includegraphics[width=0.13\linewidth]{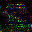}
		\hspace{0.cm}
		\includegraphics[width=0.13\linewidth]{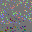}
		\hspace{0.cm}
		\includegraphics[width=0.13\linewidth]{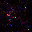}
		\hspace{0.cm}
		\includegraphics[width=0.13\linewidth]{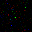}
		\hspace{0.cm}
		\includegraphics[width=0.13\linewidth]{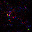}
		\\\vspace{0.15cm}
		\includegraphics[width=0.13\linewidth]{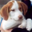}
		\hspace{0.0cm}
		\includegraphics[width=0.13\linewidth]{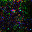}
		\hspace{0.cm}
		\includegraphics[width=0.13\linewidth]{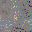}
		\hspace{0.cm}
		\includegraphics[width=0.13\linewidth]{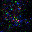}
		\hspace{0.cm}
		\includegraphics[width=0.13\linewidth]{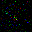}
		\hspace{0.cm}
		\includegraphics[width=0.13\linewidth]{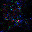}
		\caption{\scriptsize Model is trained with images with drop rates chosen randomly in $[0,1]$.}\label{fig:interp1}
	\end{subfigure}\\\vspace{0.2cm}
	\begin{subfigure}[t]{0.5\textwidth}
		\centering
		\includegraphics[width=0.13\linewidth]{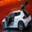}
		\hspace{0.0cm}
		\includegraphics[width=0.13\linewidth]{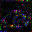}
		\hspace{0.cm}
		\includegraphics[width=0.13\linewidth]{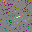}
		\hspace{0.cm}
		\includegraphics[width=0.13\linewidth]{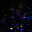}
		\hspace{0.cm}
		\includegraphics[width=0.13\linewidth]{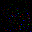}
		\hspace{0.cm}
		\includegraphics[width=0.13\linewidth]{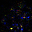}
		\\\vspace{0.15cm}
		\includegraphics[width=0.13\linewidth]{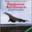}
		\hspace{0.0cm}
		\includegraphics[width=0.13\linewidth]{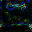}
		\hspace{0.cm}
		\includegraphics[width=0.13\linewidth]{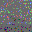}
		\hspace{0.cm}
		\includegraphics[width=0.13\linewidth]{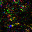}
		\hspace{0.cm}
		\includegraphics[width=0.13\linewidth]{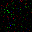}
		\hspace{0.cm}
		\includegraphics[width=0.13\linewidth]{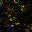}
		\\\vspace{0.15cm}
		\includegraphics[width=0.13\linewidth]{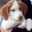}
		\hspace{0.0cm}
		\includegraphics[width=0.13\linewidth]{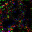}
		\hspace{0.cm}
		\includegraphics[width=0.13\linewidth]{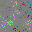}
		\hspace{0.cm}
		\includegraphics[width=0.13\linewidth]{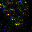}
		\hspace{0.cm}
		\includegraphics[width=0.13\linewidth]{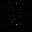}
		\hspace{0.cm}
		\includegraphics[width=0.13\linewidth]{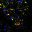}
		\caption{\small Model is trained to classify subsampled images with drop rate of $90\%$ into their true label, while mapping original images to uniform distribution. Accuracy on subsampled images is $78.9\%$.}\label{fig:interp2}
	\end{subfigure}	\\\vspace{0.2cm}
	\begin{subfigure}[t]{0.5\textwidth}
		\centering
		\includegraphics[width=0.13\linewidth]{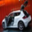}
		\hspace{0.0cm}
		\includegraphics[width=0.13\linewidth]{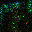}
		\hspace{0.cm}
		\includegraphics[width=0.13\linewidth]{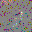}
		\hspace{0.cm}
		\includegraphics[width=0.13\linewidth]{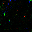}
		\hspace{0.cm}
		\includegraphics[width=0.13\linewidth]{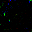}
		\hspace{0.cm}
		\includegraphics[width=0.13\linewidth]{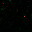}
		\\\vspace{0.15cm}
		\includegraphics[width=0.13\linewidth]{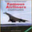}
		\hspace{0.0cm}
		\includegraphics[width=0.13\linewidth]{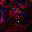}
		\hspace{0.cm}
		\includegraphics[width=0.13\linewidth]{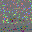}
		\hspace{0.cm}
		\includegraphics[width=0.13\linewidth]{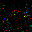}
		\hspace{0.cm}
		\includegraphics[width=0.13\linewidth]{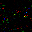}
		\hspace{0.cm}
		\includegraphics[width=0.13\linewidth]{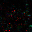}
		\\\vspace{0.15cm}
		\includegraphics[width=0.13\linewidth]{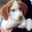}
		\hspace{0.0cm}
		\includegraphics[width=0.13\linewidth]{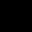}
		\hspace{0.cm}
		\includegraphics[width=0.13\linewidth]{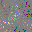}
		\hspace{0.cm}
		\includegraphics[width=0.13\linewidth]{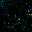}
		\hspace{0.cm}
		\includegraphics[width=0.13\linewidth]{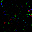}
		\hspace{0.cm}
		\includegraphics[width=0.13\linewidth]{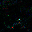}
		\caption{\small Model is trained to classify subsampled images into their true label, while mapping subsampled noisy images to uniform distribution. The subsampled noisy images are obtained as $X'=(X+\epsilon V)\circ M$, where $M$ is sampling mask, $V$ is a random variable that takes values of $\{-1,1\}$ with equal probability and $\epsilon=16/255$. The drop rate is $90\%$. Accuracy on subsampled images is $80.9\%$.}\label{fig:interp3}
	\end{subfigure}	
	\caption{\small Visualizing explanation maps. Notations: $X$, $X'$ and $M$ are original and subsampled images and the sampling mask, respectively. $E(X)$ is the explanation map on $X$, computed as $E(X)=|\partial F(X)/\partial X|$, and $E(X')$ is the explanation on $X'$. The gradient quantifies the {\it sensitivity} of model output with respect to its input. It, however, does not quantify how much each input dimension {\it contributes} to model prediction.}\label{fig:interp}
\end{figure}

We examine the interpretability for a ResNet-110 network trained with subsampled CIFAR10 images with drop rates chosen randomly in $[0,1]$. Figure~\ref{fig:interp1} shows explanation maps $E(X)$ and $E(X')$ for original and subsampled images, respectively. 
For original images, the explanation map is similar to the pattern of edges in image, a phenomenon that~\cite{adebayo2018sanity} also observed and posed as a shortcoming of interpretability methods. 
For subsampled images, however, the explanation is not informative. We visualize $E(X')\circ M$ and $E(X')\circ (1-M)$, which respectively show the gradient magnitude at pixels that have been dropped and those that are not dropped. 
As can be seen, most of larger values of gradient are at positions of dropped pixels, i.e., pixels that do not contribute to the model prediction.

The results raise questions about the usability of such techniques in explaining model predictions. 
The gradient captures the {\it sensitivity} of the model output with respect to its input, i.e., it quantifies how much a change in a small neighborhood around the input would change the predictions $F(X)$. It, however, does not quantify how much each input dimension {\it contributes} to the model prediction. Specifically, in our case, such interpretability methods do not {\it explain how} the model recognizes the image from few pixels.

For classifying subsampled images, the network might implicitly rely on features of original images, i.e., it might have learned to produce similar representations for original and subsampled images. 
In order to prevent the model to do so, we train a model to classify subsampled images into their true label, while mapping original images to uniform distribution. This training approach results in a network with accuracy of $78.9\%$ on subsampled images (with $90\%$ drop rate), which is only about $2\%$ less than a model that is only trained with subsampled images. 
The results imply that the network is capable of classifying subsampled images without actually learning features of natural images. 
Figure~\ref{fig:interp2} shows the explanation maps for few images. Similar to~\ref{fig:interp1}, the explanations do not provide insights into the model workflow.

Finally, we train a model to classify subsampled images and subsampled noisy images differently to investigate to what extent the network relies on the exact values of subsampled pixels. Specifically, we train the model to classify subsampled images into their true label, while mapping subsampled noisy images to uniform distribution. The subsampled noisy images are obtained as $X'=(X+\epsilon V)\circ M$, where $M$ is sampling mask, $V$ is a random variable that takes values of $\{-1,1\}$ with equal probability and $\epsilon=16/255$. The drop rate is $90\%$. 

Interestingly, the trained model achieves accuracy of $80.9\%$ on subsampled images, which is almost the same as a model trained only with subsampled images. 
Figure~\ref{fig:interp3} shows the explanation maps for few images. 
For this model, explanations on original images are not correlated with edge pattern. Also, explanations on subsampled images are sparser compared to~\ref{fig:interp1} and~\ref{fig:interp2}. Moreover, most of larger values of gradient are at positions where pixels have not been dropped. 
Further exploring interpretability of networks trained with subsampled images is left for future work.  


\subsection{Visualizing Convolutional Filters}

\begin{figure}[t]
	\centering
	\begin{subfigure}[t]{0.5\textwidth}
		\centering
		\includegraphics[width=0.095\linewidth]{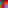}
		\hspace{0.0cm}
		\includegraphics[width=0.095\linewidth]{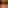}
		\hspace{0.cm}
		\includegraphics[width=0.095\linewidth]{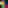}
		\hspace{0.cm}
		\includegraphics[width=0.095\linewidth]{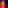}
		\hspace{0.cm}
		\includegraphics[width=0.095\linewidth]{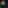}
		\hspace{0.cm}
		\includegraphics[width=0.095\linewidth]{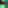}
		\hspace{0.cm}
		\includegraphics[width=0.095\linewidth]{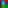}
		\hspace{0.cm}
		\includegraphics[width=0.095\linewidth]{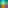}
	\end{subfigure}\\\vspace{0.15cm}
	\begin{subfigure}[t]{0.5\textwidth}
		\centering
		\includegraphics[width=0.095\linewidth]{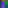}
		\hspace{0.cm}
		\includegraphics[width=0.095\linewidth]{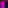}
		\hspace{0.cm}
		\includegraphics[width=0.095\linewidth]{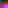}
		\hspace{0.cm}
		\includegraphics[width=0.095\linewidth]{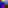}
		\hspace{0.cm}
		\includegraphics[width=0.095\linewidth]{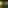}
		\hspace{0.cm}
		\includegraphics[width=0.095\linewidth]{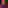}
		\hspace{0.cm}
		\includegraphics[width=0.095\linewidth]{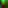}
		\hspace{0.cm}
		\includegraphics[width=0.095\linewidth]{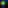}
		\caption{\small Model is trained normally.}
	\end{subfigure}\\\vspace{0.25cm}
	\begin{subfigure}[t]{0.5\textwidth}
		\centering
		\includegraphics[width=0.095\linewidth]{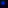}
		\hspace{0.cm}
		\includegraphics[width=0.095\linewidth]{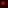}
		\hspace{0.cm}
		\includegraphics[width=0.095\linewidth]{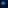}
		\hspace{0.cm}
		\includegraphics[width=0.095\linewidth]{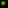}
		\hspace{0.cm}
		\includegraphics[width=0.095\linewidth]{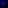}
		\hspace{0.cm}
		\includegraphics[width=0.095\linewidth]{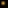}
		\hspace{0.cm}
		\includegraphics[width=0.095\linewidth]{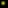}
		\hspace{0.cm}
		\includegraphics[width=0.095\linewidth]{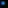}
	\end{subfigure}\\\vspace{0.15cm}
	\begin{subfigure}[t]{0.5\textwidth}
		\centering
		\includegraphics[width=0.095\linewidth]{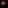}
		\hspace{0.cm}
		\includegraphics[width=0.095\linewidth]{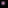}
		\hspace{0.cm}
		\includegraphics[width=0.095\linewidth]{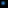}
		\hspace{0.cm}
		\includegraphics[width=0.095\linewidth]{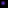}
		\hspace{0.cm}
		\includegraphics[width=0.095\linewidth]{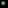}
		\hspace{0.cm}
		\includegraphics[width=0.095\linewidth]{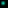}
		\hspace{0.cm}
		\includegraphics[width=0.095\linewidth]{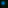}
		\hspace{0.cm}
		\includegraphics[width=0.095\linewidth]{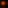}
		\caption{\small Model is trained with images with drop rate of $90\%$.}
	\end{subfigure}\\\vspace{0.25cm}
	\begin{subfigure}[t]{0.5\textwidth}
		\centering
		\includegraphics[width=0.095\linewidth]{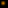}
		\hspace{0.cm}
		\includegraphics[width=0.095\linewidth]{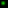}
		\hspace{0.cm}
		\includegraphics[width=0.095\linewidth]{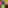}
		\hspace{0.cm}
		\includegraphics[width=0.095\linewidth]{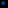}
		\hspace{0.cm}
		\includegraphics[width=0.095\linewidth]{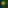}
		\hspace{0.cm}
		\includegraphics[width=0.095\linewidth]{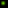}
		\hspace{0.cm}
		\includegraphics[width=0.095\linewidth]{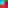}
		\hspace{0.cm}
		\includegraphics[width=0.095\linewidth]{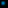}
	\end{subfigure}\\\vspace{0.15cm}
	\begin{subfigure}[t]{0.5\textwidth}
		\centering
		\includegraphics[width=0.095\linewidth]{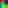}
		\hspace{0.cm}
		\includegraphics[width=0.095\linewidth]{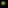}
		\hspace{0.cm}
		\includegraphics[width=0.095\linewidth]{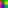}
		\hspace{0.cm}
		\includegraphics[width=0.095\linewidth]{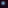}
		\hspace{0.cm}
		\includegraphics[width=0.095\linewidth]{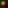}
		\hspace{0.cm}
		\includegraphics[width=0.095\linewidth]{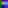}
		\hspace{0.cm}
		\includegraphics[width=0.095\linewidth]{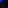}
		\hspace{0.cm}
		\includegraphics[width=0.095\linewidth]{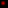}
		\caption{\scriptsize Model is trained with images with drop rates chosen randomly in $[0,1]$.}
	\end{subfigure}	
	\caption{\small Visualizing convolutional filters of first layer of ResNet-110 networks trained with CIFAR10 dataset. Models trained with subsampled images have more concentrated weights.}\label{fig:vis}
\end{figure}

Convolutional networks are known to learn basic image patterns such as edges and blobs in early layers and then combine them in later layers to distinguish  complex objects~\cite{zeiler2014visualizing}. Dropping pixels at a high rate disrupts such basic patterns. As a result, the network will not be able to readily extract spatial features of the image data. To gain insight into how the model classifies inputs, we examine convolutional filters of the first layer. Figure~\ref{fig:vis} shows the visualization of filters of ResNet-110 networks trained with CIFAR10 dataset. 
We consider three cases of a normally trained model, a model trained with subsampled images with drop rate of $90\%$ and a model trained with subsampled images with drop rates chosen randomly in $[0,1]$.

As can be seen, the model that is trained with subsampled images only has filters with large values at center position. This means that the network recognizes that there is no spatial correlation between adjacent pixels and, hence, just passes several scaled versions of the image to the next layer. 
Also, the model that is trained with subsampled images with different drop rates contains a mix of concentrated filters and other filters similar to the normally trained model.

Interestingly, having concentrated filters in first layer is also observed in adversarially trained networks on MNIST dataset, where models were trained with adversarial examples with bounded $L_{\infty}$ perturbation~\cite{madry2017towards}. 
The similar behavior of the models trained with the two approaches suggests that randomly dropping pixels is indeed related to the notion of robust features observed in adversarial training. Further exploring this relationship is left for the future work.

\begin{figure*}[t]
	\centering
	\begin{subfigure}[t]{0.35\textwidth}
		\centering
		\includegraphics[width=1\linewidth]{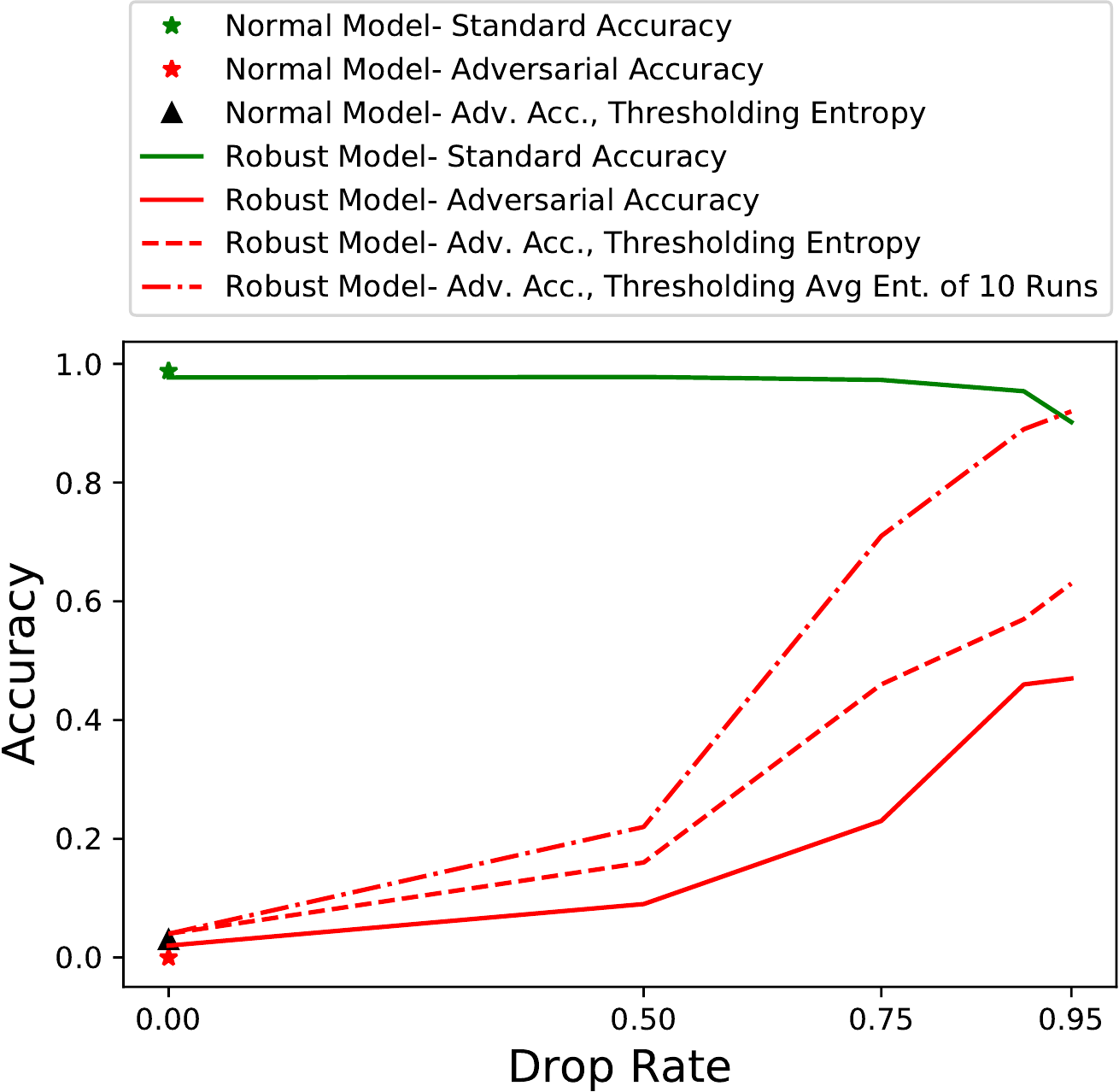}
	\end{subfigure}\hspace{0.6cm}
	\begin{subfigure}[t]{0.35\textwidth}
		\centering
		\includegraphics[width=1\linewidth]{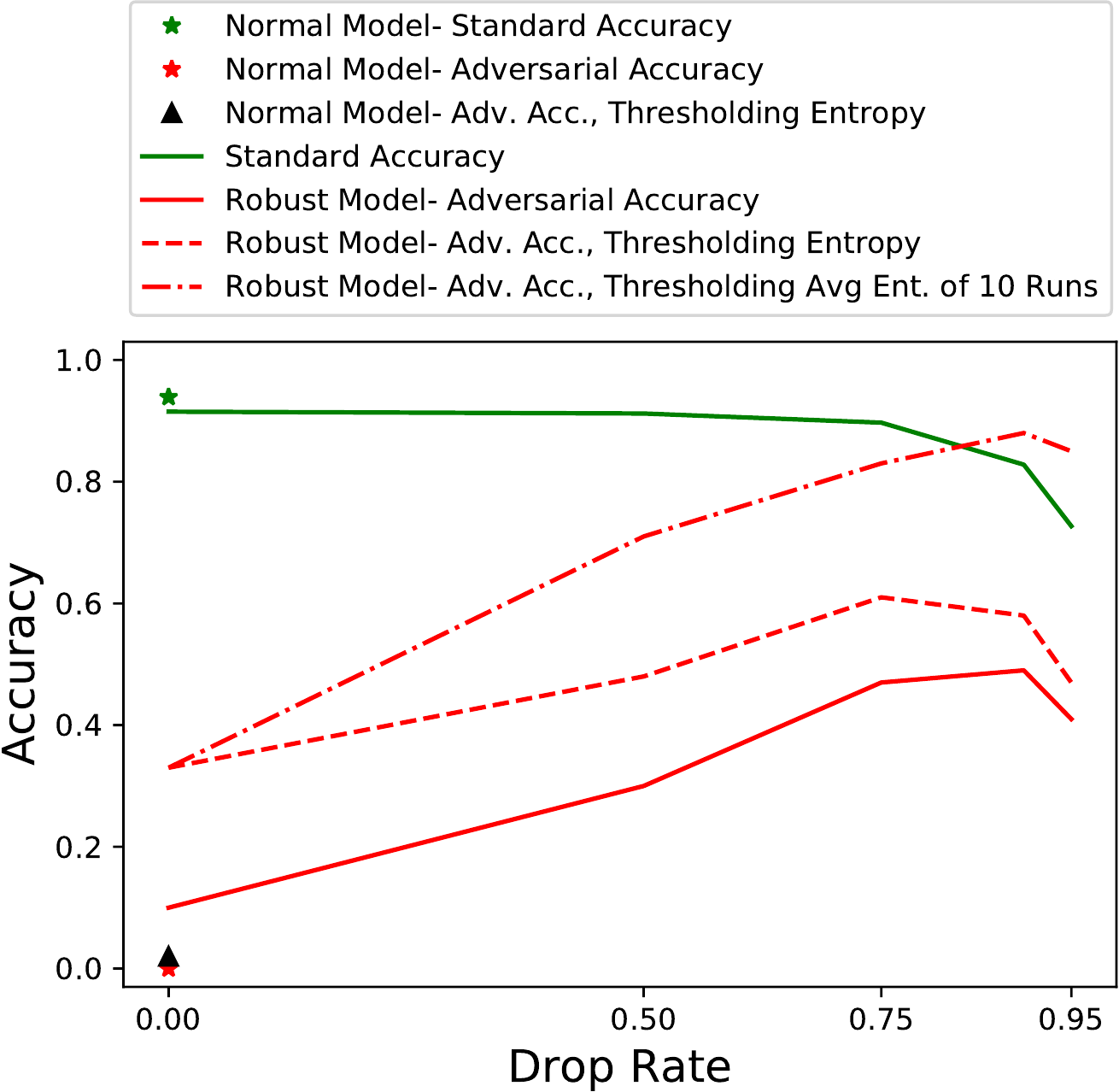}
	\end{subfigure}
	\caption{\small Accuracy on adversarial examples of {\bf (left)} GTSRB and {\bf (right)} CIFAR10 datasets with bounded $L_0$ perturbation of $0.03 \hspace{0.05cm} D$, where $D$ is the total number of  pixels. The threshold of entropy is chosen such that accuracy on clean validation images is $1\%$ less than the case where each example is tested only once.}
	\label{fig:L0}
\end{figure*}

\section{Robustness to Adversarial Examples}

In this section, we first provide a background on adversarial examples and then evaluate the robustness of models trained with subsampled images. 

\subsection{Background on Adversarial Examples}

We consider a class of adversarial examples for image classifiers where small (imperceptible) perturbation is added to an image so that the model misclassifies it (misclassification attack) or classifies it into the attacker's desired label (targeted attack). 
The perturbation is typically quantified according to an $L_p$ norm. 
The attacker's problem is formally stated as follows:
\begin{align}\label{eq:attack} 
\min &\|X'-X\|_p ,  \\
\nonumber \mbox{ s.t. } &f(X') \neq y \mbox{ or } f(X') = y_t,
\end{align}
where $X$ and $X'$ are the clean and adversarial examples, respectively, $y$ is the true label and $y_t\neq y$ is the attacker's desired target label.

We generate adversarial examples using the Projected Gradient Descent (PGD) method~\cite{kurakin2016adversarial,madry2017towards}, such that the added perturbation is bounded within $\epsilon_p$ for $L_p$ norm, i.e., $\|X'-X\|_p\leq \epsilon_p$. PGD is an iterative attack with the following update step:
\begin{align}
X^{j+1}=\Pi_{X+\mathcal{S}}(X^j + V^j), \mbox{ s.t. } \|V^j\|_p\leq \epsilon'_p,
\end{align}
where $X^j$ is the image at step $j$, $V^j$ is the attack vector at step $j$, $\epsilon'_p$ is the added perturbation per step, and $\Pi_{X+\mathcal{S}}$ is the projection operator where $\mathcal{S}$ is the set of allowed perturbations. 
According to attack goal, the attack vector is specified as follows:
\begin{itemize}
	\item $V^j = \nabla_X\ell(X^j,y)$, for misclassification attack,
	\item $V^j = -\nabla_X\ell(X^j,y_t)$, for targeted attack. 
\end{itemize}

\vspace{0.1cm}
\noindent{\bf Attack Setup.} 
To attack a model with random subsampling, we generate $10$ randomly subsampled images, for each one compute the gradient at pixels which have not been dropped, and then take average of gradients. Let $M$ be the sampling mask. The average gradient is formally obtained as follows: 
$$\Delta = \mathbb{E}_{M} [\nabla_X\ell(X \circ M,y) \circ M].$$

We consider the cases where the $L_0$, $L_2$ or $L_{\infty}$ norm of perturbation is bounded. 
For GTSRB, we set $\epsilon_0=0.03\hspace{0.1cm}D$, where $D$ is the total number of pixels, $\epsilon_2=512/255$ and $\epsilon_{\infty}=32/255$. 
The attack step size is set to $\epsilon'_0=1$, $\epsilon'_2=32/255$ and $\epsilon'_{\infty}=4/255$. 
For CIFAR10, we set $\epsilon_0=0.03\hspace{0.1cm}D$, $\epsilon_2=1$ and $\epsilon_{\infty}=16/255$. 
The attack step size is also set to $\epsilon'_0=1$, $\epsilon'_2=16/255$ and $\epsilon'_{\infty}=2/255$. 
We perform PGD attack with cross-entropy and CW~\cite{carlini2017towards} loss functions and for misclassification and targeted attacks, and present the best attack results. 

\subsection{Case of Bounded $L_0$ Perturbation}

Training with random subsampling is suited to defend against $L_0$ adversarial examples, since the model is trained to be robust to missing pixels. We also observed that subsampled adversarial examples result in more distributed output probability vector than subsampled clean images. Therefore, we  enhance the defense mechanism by rejecting examples for which the entropy of probability vector is larger than a threshold. The threshold is chosen so as to have $1\%$ false positive rate on validation data, i.e., the accuracy on clean validation images is reduced by $1\%$. 

With random subsampling, the accuracy can be improved by averaging the output on multiple different subsampled versions of the input. 
To improve adversarial robustness, we compute the average output probability vector over $10$ different subsampled inputs and reject the example if the entropy of probability vector is larger than a threshold. The threshold is chosen such that accuracy on clean validation images is $1\%$ less than the case where each example is run only once. 

Figure~\ref{fig:L0} shows the results on GTSRB and CIFAR10 datasets. As expected, larger drop rate improves adversarial robustness at the cost of reducing standard accuracy. The experiments are performed on a single model which has been trained with images with drop rates in $[0,1]$. Such a model has the advantage that at test time the drop rate can be tuned to achieve different levels of trade-offs between accuracy and robustness. 

In Figure~\ref{fig:L0}, it can be also seen that rejecting inputs based on the value of entropy improves the trade-off of accuracy and robustness, since the adversarial accuracy is increased by larger value than the $1\%$ reduction in standard accuracy. 
Moreover, averaging over $10$ runs and rejecting inputs with high entropy results in adversarial accuracy to be on par with standard accuracy at high drop rates.

\subsection{Case of Bounded $L_2$ and $L_{\infty}$ Perturbation}

\begin{figure}[t]
	\centering
	\begin{subfigure}[t]{0.23\textwidth}
		\centering
		\includegraphics[width=1\linewidth]{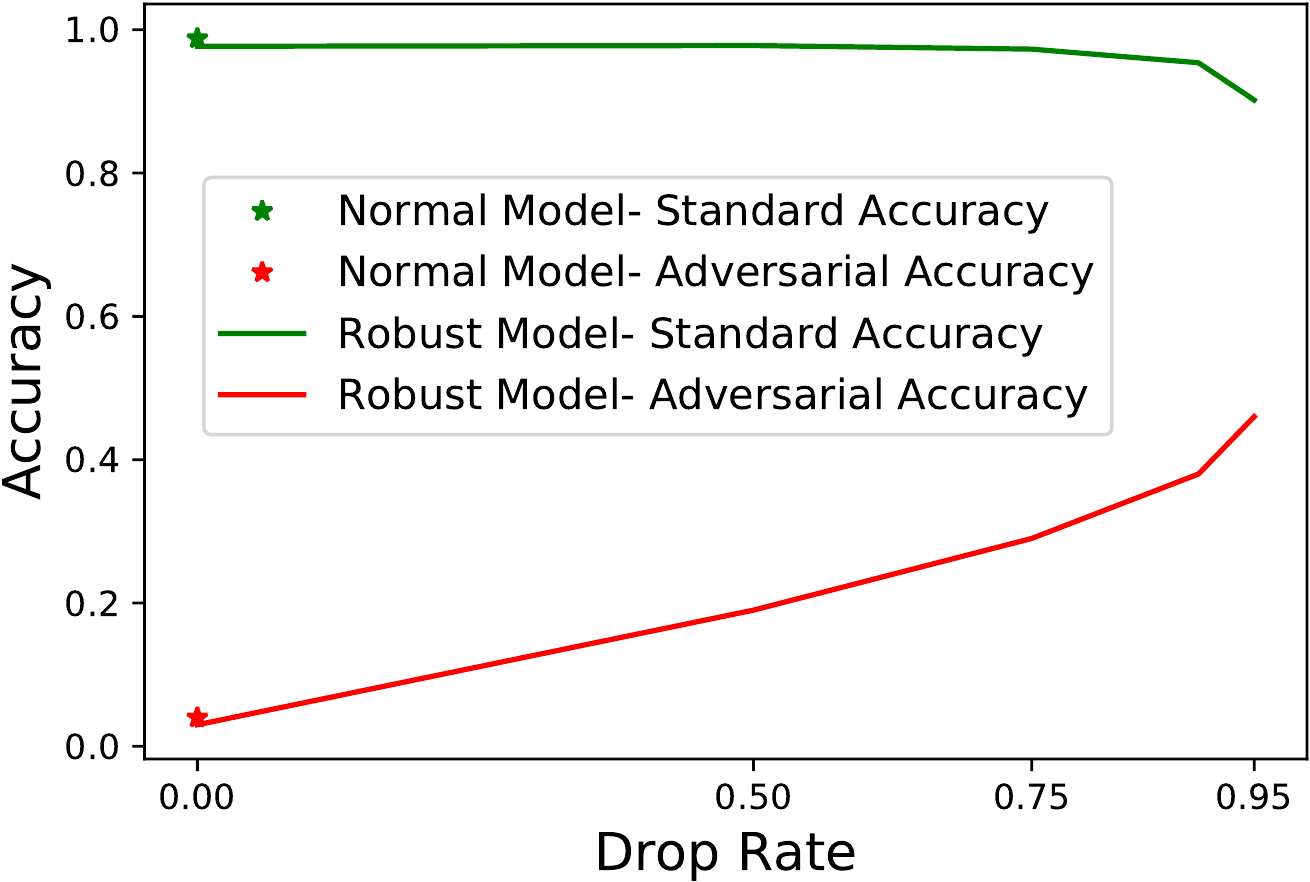}
		\caption{\small GTSRB, $L_2$ Attack.}
	\end{subfigure}\hspace{0.2cm}
	\begin{subfigure}[t]{0.23\textwidth}
		\centering
		\includegraphics[width=1\linewidth]{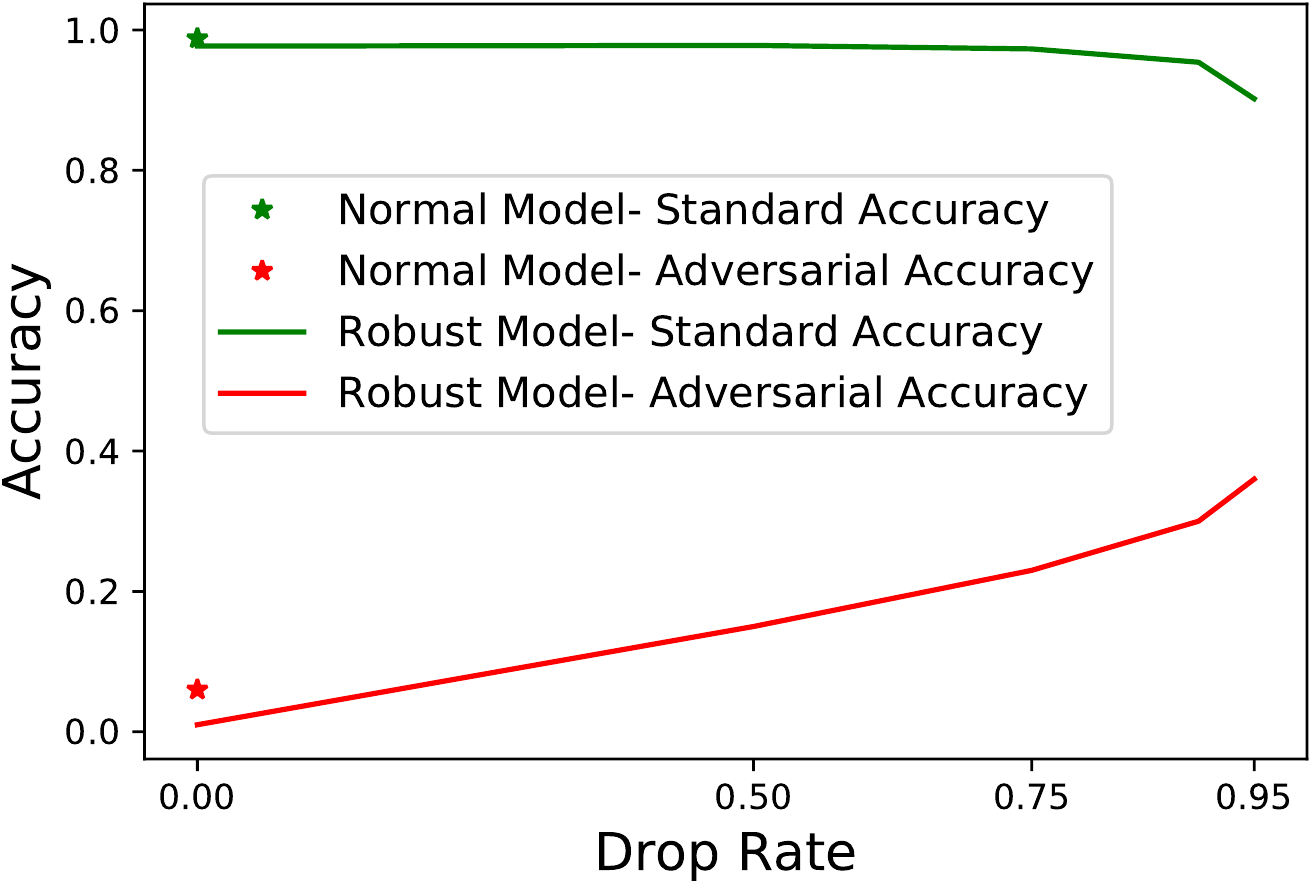}
		\caption{\small GTSRB, $L_{\infty}$ Attack.}
	\end{subfigure}\\\vspace{0.2cm}
	\begin{subfigure}[t]{0.23\textwidth}
		\centering
		\includegraphics[width=1\linewidth]{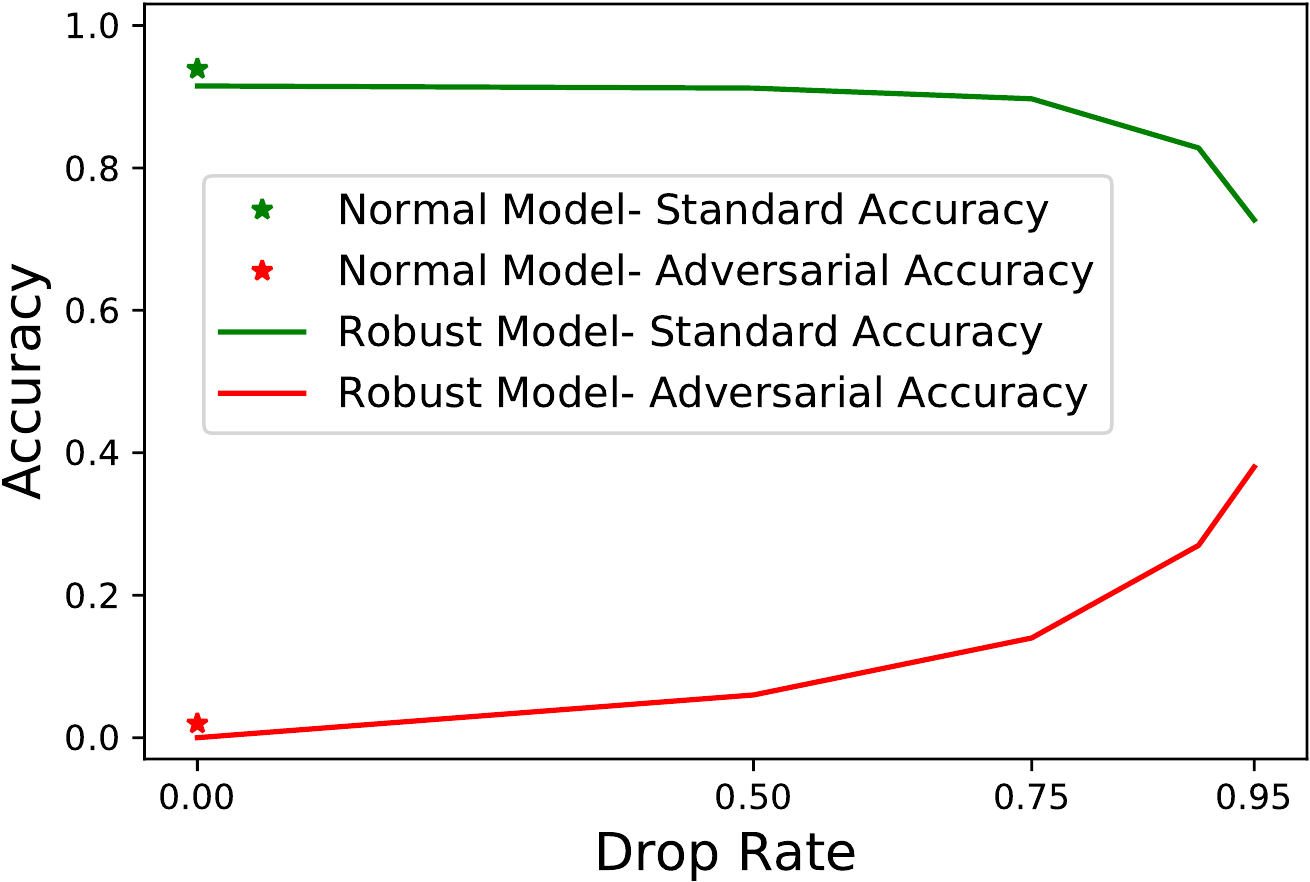}
		\caption{\small CIFAR10, $L_2$ Attack.}
	\end{subfigure}\hspace{0.2cm}
	\begin{subfigure}[t]{0.23\textwidth}
		\centering
		\includegraphics[width=1\linewidth]{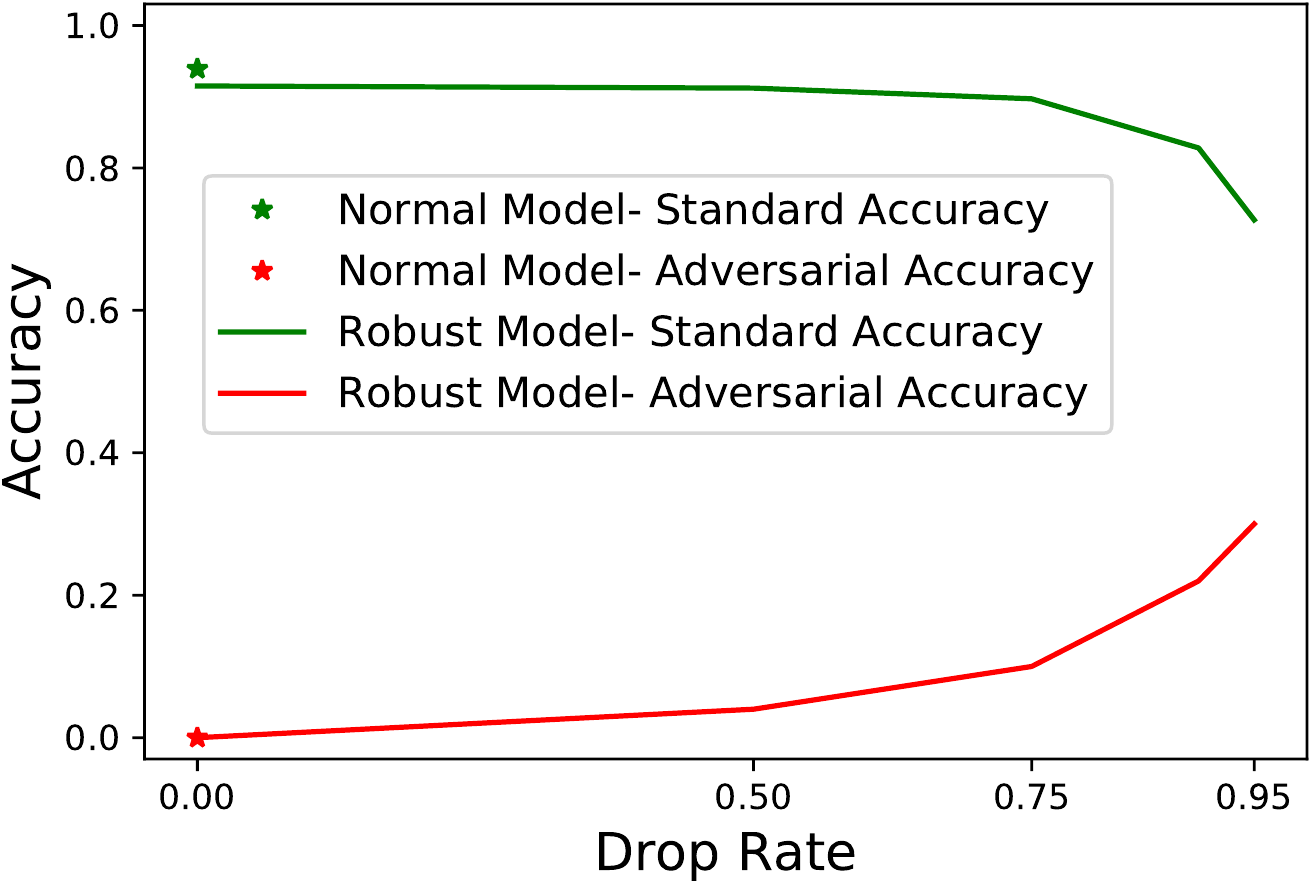}
		\caption{\small CIFAR10, $L_{\infty}$ Attack.}
	\end{subfigure}
	\caption{\small Accuracy on adversarial examples of GTSRB and CIFAR10 datasets with bounded $L_2$ and $L_{\infty}$ perturbation.}
	\label{fig:Linf}
\end{figure}

Figure~\ref{fig:Linf} shows the attack results for $L_2$ and $L_{\infty}$ cases for GTSRB and CIFAR10 datasets. 
As can be seen, random subsampling improves robustness in both cases and, similar to $L_0$ case, larger drop rate results in higher adversarial robustness. 
Intuitively, for attacking the model, the adversary needs to distribute their budget to all features in such a way that the expected attack success rate is maximized. 
This will reduce the attack effectiveness compared to the case that the attacker knows the exact sampling pattern. 




\section{Related Work}

Defenses against adversarial examples with bounded $L_p$ perturbation have been widely studied~\cite{goodfellow2014explaining,carlini2017towards,carlini2017adversarial, madry2017towards,athalye2018obfuscated,uesato2018adversarial}. Adversarial training is the state-of-the-art approach for $L_2$ and $L_{\infty}$ cases, but is shown to significantly slow down the training procedure~\cite{madry2017towards, tsipras2018there}.   
While most papers studied defenses in $L_{\infty}$ setting, some of real-world attacks based on adversarial examples fit in the $L_0$ setting~\cite{evtimov2017robust,sharif2016accessorize}. For example,~\cite{evtimov2017robust} attacked traffic sign detection algorithms by adding sticker-like perturbation to images. Also,~\cite{sharif2016accessorize} showed face recognition algorithms can be fooled by adding physically-realizable perturbation such as eyeglasses to images. 

In~\cite{bafna2018thwarting}, the authors proposed a method for improving robustness in $L_0$ attack setting by exploiting the sparsity of natural images in Fourier domain. 
They showed attack results on MNIST 
and Fashion-MNIST 
datasets and mentioned that the sparsity property might not hold in large images. Similar to their method, we used a property of natural images, namely high spatial correlation, to mitigate the effect of adversarial perturbation. Our approach is, however, general to natural images of any size. In fact, with larger images, it is possible to drop pixels at a higher rate and still restore the image~\cite{elad2005simultaneous,hosseini2013fast}. Hence, the classifier might be able to recognize subsampled images with higher drop rates and, as a result, achieve better robustness. 
Moreover, our method improves the robustness against $L_2$ and $L_{\infty}$ adversarial examples in addition to the $L_0$ case. 
As a future work, we will implement our method on Imagenet dataset.

Several papers have proposed using post-processing algorithms to increase adversarial robustness~\cite{xie2017mitigating,dhillon2018stochastic}. 
In~\cite{xie2017mitigating}, the authors proposed applying random resizing and padding at inference time. 
\cite{dhillon2018stochastic} presented an algorithm for pruning a random subset of activations of a pretrained networks and scaling up the rest. 
Unlike our method, such algorithms do not train the model to learn the randomness.

Introducing randomness to inputs or the network itself at both training and test times is recently explored and shown to improve performance on adversarial examples~\cite{liu2018towards,lecuyer2018certified}. 
In~\cite{liu2018towards}, the authors proposed adding random noise layers to the network and ensembling the prediction over random noises. 
\cite{lecuyer2018certified} adopted similar idea and used differential privacy to provide certified robustness against adversarial perturbations. 
In this paper, we proposed to train the model with subsampled images, with the drop rates randomly chosen in $[0,1]$, and test it with subsampled images with high drop rates. We showed that our method improves adversarial robustness in all cases of $L_0$, $L_2$ and $L_{\infty}$ perturbations.


\section{Conclusion}

In this paper, we showed that image classifiers can be trained to recognize images with high drop rates. We then proposed to train models with subsampled images with drop rates randomly chosen in $[0,1]$. Our experimental results on GTSRB and CIFR10 datasets showed that such models improve the robustness against adversarial examples in all cases of $L_0$, $L_2$ and $L_{\infty}$ perturbation, while reducing standard accuracy by a small value.


\vspace{0.35cm}
\noindent{\bf \large Acknowledgments}

\vspace{0.1cm}
\noindent This work was supported by ONR grant N00014-17-S-B001.


{\small
\bibliographystyle{ieeetr}
\bibliography{negbib}

\begin{thebibliography}{10}

\bibitem{Goodfellow2017openai}
I.~Goodfellow, N.~Papernot, S.~Huang, Y.~Duan, and P.~Abbeel, ``Attacking
  machine learning with adversarial examples,'' {\em Open AI Blog}, 2017.
\newblock \url{https://blog.openai.com/adversarial-example-research/}.

\bibitem{biggio2013evasion}
B.~Biggio, I.~Corona, D.~Maiorca, B.~Nelson, N.~{\v{S}}rndi{\'c}, P.~Laskov,
  G.~Giacinto, and F.~Roli, ``Evasion attacks against machine learning at test
  time,'' in {\em Joint European conference on machine learning and knowledge
  discovery in databases}, pp.~387--402, Springer, 2013.

\bibitem{szegedy2013intriguing}
C.~Szegedy, W.~Zaremba, I.~Sutskever, J.~Bruna, D.~Erhan, I.~Goodfellow, and
  R.~Fergus, ``Intriguing properties of neural networks,'' {\em arXiv preprint
  arXiv:1312.6199}, 2013.

\bibitem{carlini2017towards}
N.~Carlini and D.~Wagner, ``Towards evaluating the robustness of neural
  networks,'' in {\em 2017 IEEE Symposium on Security and Privacy (SP)},
  pp.~39--57, IEEE, 2017.

\bibitem{carlini2017adversarial}
N.~Carlini and D.~Wagner, ``Adversarial examples are not easily detected:
  Bypassing ten detection methods,'' in {\em Proceedings of the 10th ACM
  Workshop on Artificial Intelligence and Security}, pp.~3--14, ACM, 2017.

\bibitem{athalye2018obfuscated}
A.~Athalye, N.~Carlini, and D.~Wagner, ``Obfuscated gradients give a false
  sense of security: Circumventing defenses to adversarial examples,'' {\em
  arXiv preprint arXiv:1802.00420}, 2018.

\bibitem{goodfellow2014explaining}
I.~J. Goodfellow, J.~Shlens, and C.~Szegedy, ``Explaining and harnessing
  adversarial examples,'' {\em arXiv preprint arXiv:1412.6572}, 2014.

\bibitem{madry2017towards}
A.~Madry, A.~Makelov, L.~Schmidt, D.~Tsipras, and A.~Vladu, ``Towards deep
  learning models resistant to adversarial attacks,'' {\em arXiv preprint
  arXiv:1706.06083}, 2017.

\bibitem{kannan2018adversarial}
H.~Kannan, A.~Kurakin, and I.~Goodfellow, ``Adversarial logit pairing,'' {\em
  arXiv preprint arXiv:1803.06373}, 2018.

\bibitem{tsipras2018there}
D.~Tsipras, S.~Santurkar, L.~Engstrom, A.~Turner, and A.~Madry, ``There is no
  free lunch in adversarial robustness (but there are unexpected benefits),''
  {\em arXiv preprint arXiv:1805.12152}, 2018.

\bibitem{elad2005simultaneous}
M.~Elad, J.-L. Starck, P.~Querre, and D.~L. Donoho, ``Simultaneous cartoon and
  texture image inpainting using morphological component analysis (mca),'' {\em
  Applied and Computational Harmonic Analysis}, vol.~19, no.~3, pp.~340--358,
  2005.

\bibitem{hosseini2013fast}
H.~Hosseini and F.~Marvasti, ``Fast restoration of natural images corrupted by
  high-density impulse noise,'' {\em EURASIP Journal on Image and Video
  Processing}, vol.~2013, no.~1, p.~15, 2013.

\bibitem{kurakin2016adversarial}
A.~Kurakin, I.~Goodfellow, and S.~Bengio, ``Adversarial machine learning at
  scale,'' {\em arXiv preprint arXiv:1611.01236}, 2016.

\bibitem{stallkamp2012man}
J.~Stallkamp, M.~Schlipsing, J.~Salmen, and C.~Igel, ``Man vs. computer:
  Benchmarking machine learning algorithms for traffic sign recognition,'' {\em
  Neural networks}, vol.~32, pp.~323--332, 2012.

\bibitem{krizhevsky2009learning}
A.~Krizhevsky and G.~Hinton, ``Learning multiple layers of features from tiny
  images,'' tech. rep., Citeseer, 2009.

\bibitem{he2016deep}
K.~He, X.~Zhang, S.~Ren, and J.~Sun, ``Deep residual learning for image
  recognition,'' in {\em Proceedings of the IEEE conference on computer vision
  and pattern recognition}, pp.~770--778, 2016.

\bibitem{simonyan2013deep}
K.~Simonyan, A.~Vedaldi, and A.~Zisserman, ``Deep inside convolutional
  networks: Visualising image classification models and saliency maps,'' {\em
  arXiv preprint arXiv:1312.6034}, 2013.

\bibitem{shrikumar2016not}
A.~Shrikumar, P.~Greenside, A.~Shcherbina, and A.~Kundaje, ``Not just a black
  box: Learning important features through propagating activation
  differences,'' {\em arXiv preprint arXiv:1605.01713}, 2016.

\bibitem{selvaraju2016grad}
R.~R. Selvaraju, A.~Das, R.~Vedantam, M.~Cogswell, D.~Parikh, and D.~Batra,
  ``Grad-cam: Why did you say that?,'' {\em arXiv preprint arXiv:1611.07450},
  2016.

\bibitem{smilkov2017smoothgrad}
D.~Smilkov, N.~Thorat, B.~Kim, F.~Vi{\'e}gas, and M.~Wattenberg, ``Smoothgrad:
  removing noise by adding noise,'' {\em arXiv preprint arXiv:1706.03825},
  2017.

\bibitem{adebayo2018sanity}
J.~Adebayo, J.~Gilmer, M.~Muelly, I.~Goodfellow, M.~Hardt, and B.~Kim, ``Sanity
  checks for saliency maps,'' in {\em Advances in Neural Information Processing
  Systems}, pp.~9525--9536, 2018.

\bibitem{zeiler2014visualizing}
M.~D. Zeiler and R.~Fergus, ``Visualizing and understanding convolutional
  networks,'' in {\em European conference on computer vision}, pp.~818--833,
  Springer, 2014.

\bibitem{uesato2018adversarial}
J.~Uesato, B.~O'Donoghue, A.~v.~d. Oord, and P.~Kohli, ``Adversarial risk and
  the dangers of evaluating against weak attacks,'' {\em arXiv preprint
  arXiv:1802.05666}, 2018.

\bibitem{evtimov2017robust}
I.~Evtimov, K.~Eykholt, E.~Fernandes, T.~Kohno, B.~Li, A.~Prakash, A.~Rahmati,
  and D.~Song, ``Robust physical-world attacks on deep learning models,'' {\em
  arXiv preprint arXiv:1707.08945}, vol.~1, p.~1, 2017.

\bibitem{sharif2016accessorize}
M.~Sharif, S.~Bhagavatula, L.~Bauer, and M.~K. Reiter, ``Accessorize to a
  crime: Real and stealthy attacks on state-of-the-art face recognition,'' in
  {\em Proceedings of the 2016 ACM SIGSAC Conference on Computer and
  Communications Security}, pp.~1528--1540, ACM, 2016.

\bibitem{bafna2018thwarting}
M.~Bafna, J.~Murtagh, and N.~Vyas, ``Thwarting adversarial examples: An $ l\_0
  $-robust sparse fourier transform,'' in {\em Advances in Neural Information
  Processing Systems}, pp.~10096--10106, 2018.

\bibitem{xie2017mitigating}
C.~Xie, J.~Wang, Z.~Zhang, Z.~Ren, and A.~Yuille, ``Mitigating adversarial
  effects through randomization,'' {\em arXiv preprint arXiv:1711.01991}, 2017.

\bibitem{dhillon2018stochastic}
G.~S. Dhillon, K.~Azizzadenesheli, Z.~C. Lipton, J.~Bernstein, J.~Kossaifi,
  A.~Khanna, and A.~Anandkumar, ``Stochastic activation pruning for robust
  adversarial defense,'' {\em arXiv preprint arXiv:1803.01442}, 2018.

\bibitem{liu2018towards}
X.~Liu, M.~Cheng, H.~Zhang, and C.-J. Hsieh, ``Towards robust neural networks
  via random self-ensemble,'' in {\em Proceedings of the European Conference on
  Computer Vision (ECCV)}, pp.~369--385, 2018.

\bibitem{lecuyer2018certified}
M.~Lecuyer, V.~Atlidakis, R.~Geambasu, D.~Hsu, and S.~Jana, ``Certified
  robustness to adversarial examples with differential privacy,'' {\em arXiv
  preprint arXiv:1802.03471}, 2018.

\end{thebibliography}
}

\end{document}